\documentclass[10pt,twocolumn,letterpaper]{article}

\usepackage{cvpr}              
\usepackage{pifont}

\usepackage{wrapfig}

\definecolor{cvprblue}{rgb}{0.21,0.49,0.74}
\usepackage[pagebackref,breaklinks,colorlinks,allcolors=cvprblue]{hyperref}

\def\paperID{*****} 
\def\confName{CVPR}
\def\confYear{2026}

\title{VGGT-HPE: Reframing Head Pose Estimation as Relative Pose Prediction}

\author{
\normalsize{
Vasiliki Vasileiou\textsuperscript{1,2,4} \quad
Panagiotis P. Filntisis\textsuperscript{2,3}\thanks{The research work of P.~P.~Filntisis and P.~Maragos was supported by the project ``Applied Research for Autonomous Robotic Systems'' (MIS 5200632), implemented within the framework of the National Recovery and Resilience Plan ``Greece 2.0'' (Measure: 16618 -- Basic and Applied Research), and funded by the European Union -- NextGenerationEU.} \quad
Petros Maragos\textsuperscript{2,3,4}\footnotemark[1] \quad
Kostas Daniilidis\textsuperscript{1,5}
}\\
\vspace{-0.2cm}\\
\footnotesize{
\textsuperscript{1}Archimedes, Athena Research Center, Marousi, Greece \quad
\textsuperscript{2}HERON -- Hellenic Robotics Center of Excellence, Athens, Greece
}\\
\footnotesize{
\textsuperscript{3}Robotics Institute, Athena Research Center, Marousi, Greece \quad
\textsuperscript{4}School of ECE, National Technical University of Athens, Greece
}\\
\footnotesize{
\textsuperscript{5}University of Pennsylvania
}
}

\begin{document}
\maketitle
\begin{abstract}
Monocular head pose estimation is traditionally formulated as direct regression from a single image to an absolute pose. This paradigm forces the network to implicitly internalize a dataset-specific canonical reference frame. In this work, we argue that predicting the relative rigid transformation between two observed head configurations is a fundamentally easier and more robust formulation. We introduce VGGT-HPE, a relative head pose estimator built upon a general-purpose geometry foundation model. Fine-tuned exclusively on synthetic facial renderings, our method sidesteps the need for an implicit anchor by reducing the problem to estimating a geometric displacement from an explicitly provided anchor with a known pose. As a practical benefit, the relative formulation also allows the anchor to be chosen at test time — for instance, a near-neutral frame or a temporally adjacent one — so that the prediction difficulty can be controlled by the application. Despite zero real-world training data, VGGT-HPE achieves state-of-the-art results on the BIWI benchmark, outperforming established absolute regression methods trained on mixed and real datasets. Through controlled easy- and hard-pair benchmarks, we also systematically validate our core hypothesis: relative prediction is intrinsically more accurate than absolute regression, with the advantage scaling alongside the difficulty of the target pose. Project page and code: \url{https://vasilikivas.github.io/VGGT-HPE}
\end{abstract}    
\section{Introduction}
\label{sec:intro}

Head pose estimation - recovering the orientation of a human head from a single image - is a fundamental building block in computer vision~\cite{murphy2009head,abate2022head}. Accurate head pose drives a wide range of downstream applications, from gaze estimation~\cite{zhang2019mpiigaze} and driver attention monitoring~\cite{jha2022driver} to sign language recognition~\cite{koller2020quantitative}, facial behavior analysis in the wild~\cite{kollias2021affect}, human-robot interaction~\cite{mavridis2015review}, and augmented reality~\cite{marchand2016pose}. Despite decades of progress, robust head pose estimation in unconstrained settings remains challenging due to appearance variability, partial occlusions, and the inherent ambiguity of monocular 3D inference.

\begin{figure}[t]
  \centering
  \includegraphics[width=0.97 \columnwidth]{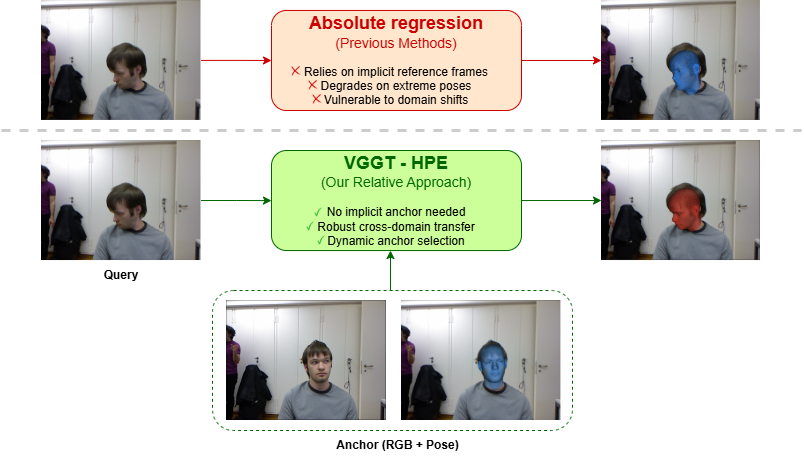}
  \caption{Absolute methods (top) regress pose from a single image relative to an implicit canonical frame learned during training, making them brittle under domain shift and extreme poses. VGGT-HPE (bottom) predicts the relative transformation between an anchor with known pose and a query, recovering the target pose by composition. This sidesteps the need for an implicit reference frame, transfers better from synthetic to real data, and allows the anchor to be chosen at test time to control prediction difficulty.}
  \label{fig:teaser}
\end{figure}

The dominant paradigm in learned head pose estimation treats the problem as direct regression from a single image to an absolute pose vector --- typically three Euler angles (yaw, pitch, roll) and, less commonly, a 3D translation. Methods such as HopeNet~\cite{ruiz2018hopenet}, 6DRepNet~\cite{hempel20226drepnet}, and WHENet~\cite{zhou2020whenet} train deep networks to map face crops directly to global orientation. This requires the model to internalize a canonical reference frame from the training data - one that is never explicitly provided at test time. More recent works improve robustness through better rotation representations~\cite{hempel20226drepnet}, transformer-based modeling~\cite{zhang2023tokenhpe}, or explicit facial geometry~\cite{chun2024trg}, but the core formulation remains unchanged. As we show in our experiments, these models tend to degrade for extreme poses, where the implicit anchor - a neutral, frontal configuration baked into the learned weights - lies far from the target.

We argue that relative head pose estimation is a fundamentally easier problem. Predicting the rigid transformation between two observed head configurations sidesteps the need for an implicit canonical frame and reduces the task to estimating a displacement. Because it observes both states, the network can solve this via implicit feature matching—shifting the paradigm from absolute classification to direct image matching. This is simpler and transfers better across domains, since relative geometry does not depend on absolute coordinate conventions or appearance distributions specific to the training set. The relative formulation also introduces a practical degree of freedom that absolute methods lack: the anchor can be chosen at test time. In video, for instance, a temporally close frame with known pose keeps the relative rotation gap small, placing the prediction well within the model's reliable operating range.

In this work we introduce VGGT-HPE, a head pose estimator built on the Visual Geometry Grounded Transformer (VGGT)~\cite{wang2025vggt}, a large-scale geometry foundation model pre-trained for general-purpose camera pose and 3D reconstruction. VGGT already learns to estimate relative camera poses between arbitrary view pairs, making it a natural backbone for our relative formulation. We fine-tune VGGT with LoRA~\cite{hu2022lora} on synthetic two-view head pairs rendered from FLAME~\cite{li2017flame}. At test time, we compose the predicted relative pose with a known anchor pose to recover the absolute target pose in any desired coordinate frame. Despite training exclusively on synthetic data, VGGT-HPE achieves state-of-the-art results on the BIWI benchmark~\cite{fanelli2013biwi} outperforming methods trained on mixed and real datasets. 

Our contributions are as follows:
\begin{itemize}
    \item A relative formulation for head pose estimation that reframes the problem as rigid displacement prediction between two views, eliminating the dependence on an implicit canonical frame and enabling strategic anchor selection at test time.
    \item A lightweight architecture that adapts VGGT~\cite{wang2025vggt}, a general-purpose geometry foundation model, to head-specific relative pose estimation via LoRA fine-tuning~\cite{hu2022lora} on synthetic FLAME~\cite{li2017flame} renderings, requiring no real-world training data.
    \item State-of-the-art results on BIWI~\cite{fanelli2013biwi} using only synthetic training data, and a detailed analysis through controlled benchmarks - easy-pair, hard-pair, and binned error-vs-rotation-gap sweeps - that directly validates the hypothesis that relative prediction is fundamentally easier than absolute regression, with the advantage growing as the anchor-target displacement increases.
\end{itemize}

\section{Related Work}
\paragraph{Monocular Head Pose Estimation.} Early head pose estimation methods relied on facial landmarks and 3D model fitting, typically recovering pose through geometric alignment or PnP-style optimization. More recent learning-based approaches instead regress pose directly from a face crop. HopeNet~\cite{ruiz2018hopenet} introduced a combined classification-and-regression strategy for Euler-angle prediction, showing that direct image-to-pose learning can be competitive without explicit landmark fitting. Subsequent works improved robustness to wider pose ranges and better rotation representations. WHENet~\cite{zhou2020whenet} targeted wide-range yaw estimation with a lightweight design suitable for real-time use, while 6DRepNet~\cite{hempel20226drepnet} replaced Euler-angle regression with a continuous 6D rotation representation and a geodesic loss, improving stability under large rotations. Transformer-based methods such as TokenHPE~\cite{zhang2023tokenhpe} further demonstrated that richer relational modeling over facial appearance can improve monocular head pose estimation. More recent 6DoF methods, including TRG~\cite{chun2024trg}, also incorporate facial geometry explicitly in order to improve translation and full pose prediction. Despite these differences, the dominant paradigm remains the same: a single image is mapped directly to an absolute pose defined in a dataset-specific canonical frame. As a result, the network must implicitly learn both the rigid orientation and the hidden reference frame relative to which that orientation is expressed. Furthermore, training these absolute regressors typically requires massive, real-world datasets often expanded via synthetic face profiling (e.g., 300W-LP~\cite{zhu2016300wlp}), which leaves them vulnerable to domain shifts. Our relative formulation, by contrast, treats pose as an image matching task, allowing it to generalize across domains even when trained exclusively on synthetic data.
\begin{figure*}[t]
  \centering
  \includegraphics[width=1.0\linewidth]{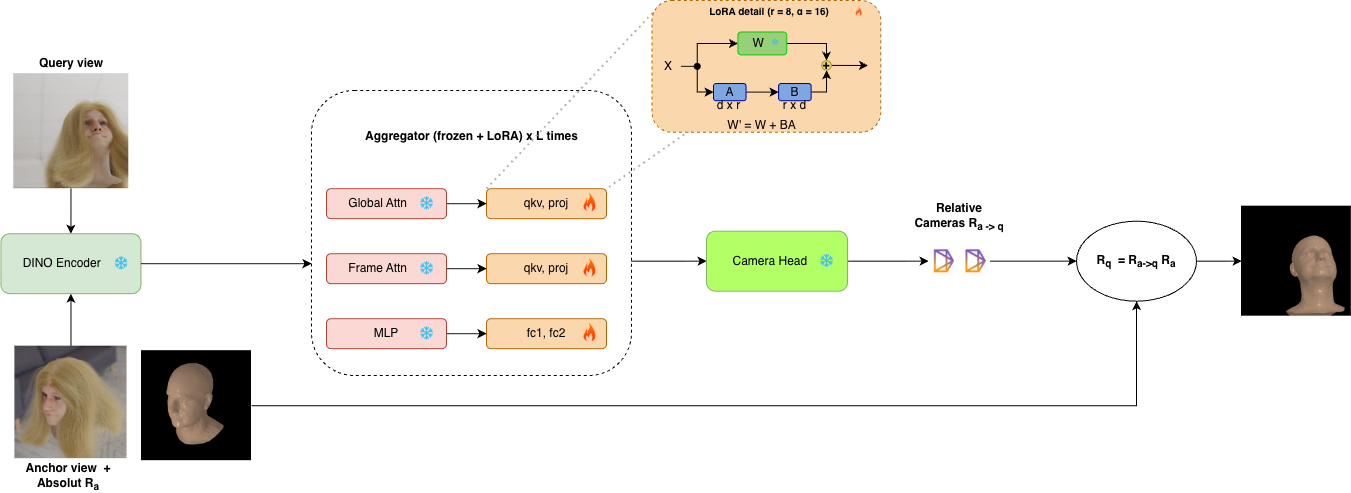}
  \caption{\textbf{Overview of the VGGT-HPE architecture.} The model takes an anchor image ($I_a$) and a query image ($I_q$) to predict the relative rigid transformation ($T_{q \leftarrow a}$) between them. We use a frozen, pre-trained VGGT backbone efficiently fine-tuned with LoRA for the facial domain. The final absolute query pose ($\hat{T}_q$) is recovered by composing the predicted relative displacement with the known anchor pose.}
  \label{fig:architecture}
\end{figure*}

\paragraph{Relative Pose Formulations and Geometric Composition.} 
Estimating the relative pose between two views is a standard formulation in visual odometry, SLAM, and rigid object tracking~\cite{engel2014lsd, wang2024dvmnet, piazza2021deep}. In these settings, predicting the relative transformation between a query and a reference view tends to be more robust and to generalize better than regressing an absolute pose in a fixed coordinate system. In head pose estimation, however, most learning-based methods still operate in the absolute, single-image regime. Temporal or multi-frame information is sometimes used in video-based head pose estimation, but typically for recurrent feature aggregation, temporal smoothing, or heuristic tracking~\cite{gu2017rnn, btc2022temporal}---not for explicit geometric composition. A few works explore delta-pose or relative rotation for specific sub-tasks, such as enforcing temporal consistency or regularizing semi-supervised training~\cite{kuhn2023relative}, but they do not output explicit relative pose transformations that can be composed at test time. Our work brings the relative formulation into head pose estimation. By predicting the relative pose between an anchor with known pose and a target frame, we remove the dependence on an implicit, dataset-specific canonical frame, connecting head pose estimation to the reference-based matching formulations used in general 3D vision.

\paragraph{Geometry Foundation Models.} 
Recent progress in 3D computer vision has moved from task-specific architectures toward large-scale geometry foundation models. CroCo~\cite{weinzaepfel2022croco}, DUSt3R~\cite{wang2024dust3r}, and MASt3R~\cite{alsfasser2024mast3r} showed that general-purpose 3D priors---dense point maps, depth, and relative camera poses---can be learned through pre-training on large, diverse multi-view datasets. VGGT~\cite{wang2025vggt} extends this line of work with a feed-forward transformer that jointly infers camera parameters and 3D scene structure from uncalibrated views. Traditional head pose estimators must learn perspective projection and 3D rotation from scratch on limited facial datasets. Models like VGGT, by contrast, already encode an understanding of epipolar geometry and relative spatial transformations from their pre-training data. Our approach builds on this: rather than designing a new architecture for faces, we adapt the pre-trained geometric reasoning of VGGT, showing that a general 3D foundation model can be efficiently specialized to head pose estimation.

\section{Methodology}

\paragraph{Overview.}
Our goal is to estimate the pose of a query head image by predicting its rigid transformation relative to an anchor image with known pose. The key intuition is that absolute methods must map a single crop to a pose defined in an implicit canonical frame that is never provided at test time, while the relative formulation turns the problem into image matching between two visible inputs --- a task that is geometrically simpler and naturally sidesteps the domain gap. We build on VGGT~\cite{wang2025vggt}, a geometry foundation model whose camera pose branch already estimates relative rigid transformations between two images, and specialize it to the facial domain via LoRA fine-tuning on synthetic FLAME renderings. An overview of our architecture can be seen in Fig.~\ref{fig:architecture}.

\paragraph{Backbone and Adaptation.}
VGGT is a feed-forward transformer pretrained on large-scale multi-view data to jointly predict camera parameters, depth, point maps, and correspondences. We retain only the camera branch and disable all other heads. Each camera pose is encoded as
\begin{equation}
p = [t, q, \phi_h, \phi_w],
\end{equation}
where $t \in \mathbb{R}^3$ is translation, $q \in \mathbb{R}^4$ is a quaternion rotation, and $\phi_h, \phi_w$ are the horizontal and vertical fields of view.

To adapt VGGT without destroying its pretrained priors, we use LoRA~\cite{hu2022lora}: for each pretrained linear layer $W$, a low-rank update $\Delta W = BA$ is learned while $W$ remains frozen, yielding $W' = W + BA$. LoRA modules are inserted into the attention and MLP projections of both the frame-level and global transformer blocks, with rank $8$ and scaling factor $16$.

\paragraph{Relative Formulation.}
Given an anchor image $I_a$ with known pose $T_a \in SE(3)$ and a query image $I_q$ with unknown pose $T_q$, the model predicts the relative transformation
\begin{equation}
T_{q \leftarrow a} = T_q \, T_a^{-1}.
\end{equation}
The absolute query pose is then recovered by composition:
\begin{equation}
\hat{T}_q = \hat{T}_{q \leftarrow a} \, T_a.
\end{equation}

The advantage over absolute regression is twofold. First, absolute methods must map a single crop to a pose defined in a canonical frame that is never provided --- it exists only implicitly in the training data. The relative formulation removes this requirement: the model just estimates how much the head moved between two images it can see. Second, by conditioning on a pair of images, the model can perform implicit feature matching between anchor and query, grounding its prediction in visual correspondence rather than memorized pose distributions. This makes the task closer to image matching than to classification, which is both geometrically simpler and naturally robust to domain shift --- the relative displacement between two views of a head looks the same whether the images are synthetic or real.

During training, we normalize each pair to the anchor frame: given extrinsics $\{T_1, T_2\}$, we set $\tilde{T}_i = T_1^{-1} T_i$ so the anchor becomes the identity. In contrast with the original VGGT~\cite{wang2025vggt}, which supervises the poses of all input frames, we supervise only the second frame's pose. We also study an absolute baseline (VGGT-HPE-Abs) using the same backbone and data but without relative normalization, to isolate the effect of the formulation.

\begin{figure}[t]
  \centering
  \includegraphics[width=1.0\linewidth, trim=0 810bp 0 0, clip]{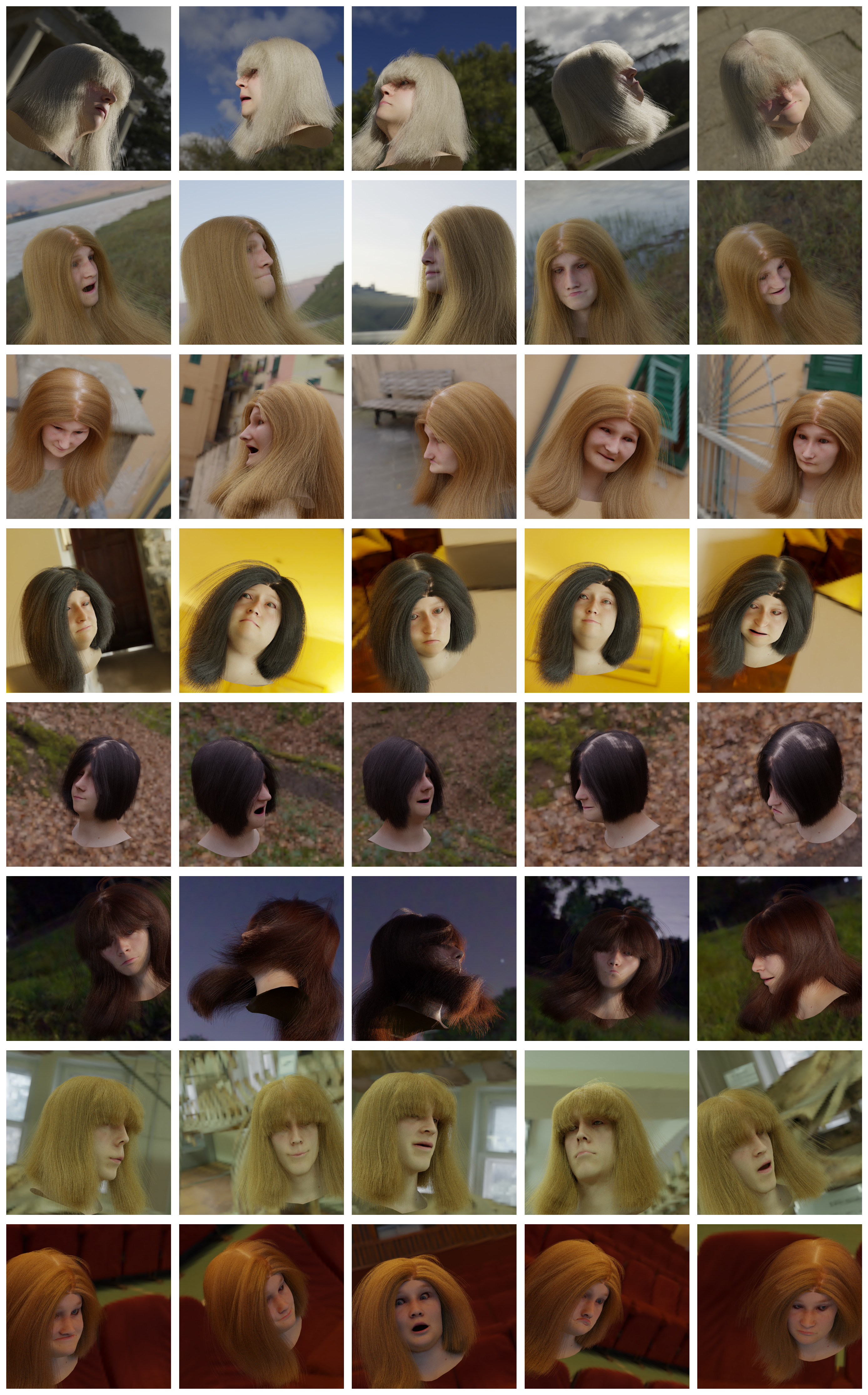}
  \caption{Samples from our synthetic training set. Each row shows a single identity rendered under varying viewpoints, expressions, and HDRI environment maps.}
  \label{fig:synth_samples}
\end{figure}

\paragraph{Synthetic Training Data.}
We generate training pairs using FLAME~\cite{li2017flame} head meshes rendered in Blender~\cite{blender2018}, following a pipeline similar to~\cite{filntisis2026mochi}. The dataset comprises 250 identities --- 200 with hairstyles from HAAR~\cite{Sklyarova2024haar} and 50 bald --- each rendered under two HDRI environment maps with ten FLAME expressions and five viewpoints per expression, yielding 25k images. Albedo is sampled from 54 textures covering diverse skin tones. Each pair shares identity and lighting while expression and viewpoint vary freely. At training time, we crop a square face region with random margin and shift, updating intrinsics accordingly, and apply pairwise appearance augmentation (color jitter, gamma, CLAHE, RGB shift, blur, noise) coherently across both views. We show representative samples of our rendered scenes in Fig.~\ref{fig:synth_samples}.

\paragraph{Training Objective.}
We supervise not only rotation but also translation and field of view, as we found that jointly predicting all camera parameters improves rotation accuracy even when only orientation is needed at test time. VGGT produces pose predictions at $K$ stages. We supervise all stages with exponentially decayed weights:
\begin{equation}
\mathcal{L}_{\mathrm{cam}}
= \frac{1}{K} \sum_{k=1}^{K} \gamma^{K-k}
\Big( \lambda_T \mathcal{L}_T^{(k)} + \lambda_R \mathcal{L}_R^{(k)} + \lambda_F \mathcal{L}_F^{(k)} \Big),
\end{equation}
where $\gamma$ is a decay factor. The translation loss is $\mathcal{L}_T = \|\hat{t} - t\|_1$ and the rotation loss operates in quaternion space: $\mathcal{L}_R = \|\hat{q} - q\|_1$.

For field-of-view supervision, we supervise the inter-frame focal ratio rather than absolute focal lengths per frame. Defining $r(\phi) = \log \tan(\phi/2)$, the focal loss is
\begin{equation}
\mathcal{L}_F = \left\| \big(\hat{r}(\phi_2) - \hat{r}(\phi_1)\big) - \big(r(\phi_2) - r(\phi_1)\big) \right\|_1.
\end{equation}
We set $\lambda_T = 1.0$, $\lambda_R = 1.0$, and $\lambda_F = 0.5$. 

\paragraph{Optimization.}
We fine-tune only the LoRA parameters with AdamW and mixed-precision training, keeping the full VGGT backbone frozen. Training runs for 200 epochs on a single A100 64\,GB GPU and completes in approximately one day.

\section{Experiments}
\label{sec:experiments}

\subsection{Experimental Setup}


\noindent\textbf{Evaluation Benchmark.}
We evaluate on the BIWI Kinect Head Pose Database~\cite{fanelli2013biwi}, which provides ground-truth 6DoF head poses for 24 subjects recorded with an RGB-D sensor. Since BIWI annotations are defined in the depth-camera coordinate frame, we transform all poses to the RGB-camera frame using the per-subject calibration parameters before evaluation. Face detection is performed with a shared MTCNN~\cite{zhang2016mtcnn} detector for all methods, following the evaluation protocol of \cite{hempel20226drepnet}.

\noindent\textbf{Baselines.}
We compare against multiple head pose estimation methods evaluated on BIWI, spanning both rotation-only and 6DoF approaches: HopeNet~\cite{ruiz2018hopenet}, FSA-Net~\cite{yang2019fsanet}, WHENet-V~\cite{zhou2020whenet}, 6DRepNet~\cite{hempel20226drepnet}, TriNet~\cite{cao2021trinet}, TokenHPE~\cite{zhang2023tokenhpe}, img2pose~\cite{albiero2021img2pose}, PerspNet~\cite{kao2023perspnet}, and TRG~\cite{chun2024trg}. All baselines are trained on real-world data or mixed regimes (e.g., real images expanded via synthetic pose profiling like 300W-LP~\cite{zhu2016300wlp}, ARKitFace~\cite{kao2023perspnet}, or combinations thereof).

\noindent\textbf{Evaluation Protocol.}
We report mean absolute error (MAE) in degrees for yaw, pitch, and roll, along with the overall MAE averaged across the three axes. For the relative model (VGGT-HPE), inference proceeds in two-view mode: an anchor frame with known ground-truth pose is paired with each target frame, the model predicts the relative rigid transformation, and the final target pose is recovered by composing the prediction with the anchor pose.

\begin{table}[t]
  \caption{Cross-domain evaluation on BIWI. Rotation errors in degrees (lower is better). Best in \textbf{bold}, second-best \underline{underlined}. Top: numbers reported in original papers (from~\cite{chun2024trg}). Bottom: reproduced under our shared MTCNN detection protocol. ``Data'' indicates the training-data regime: synthetic-only (S), real-only (R), or mixed (M; real images combined with synthetic augmentation / synthetic pose expansion, e.g., 300W-LP).}
  \label{tab:biwi_cross_domain}
  \centering
  \resizebox{\columnwidth}{!}{%
  \begin{tabular}{@{}lccccc@{}}
    \toprule
    Method & Yaw $\downarrow$ & Pitch $\downarrow$ & Roll $\downarrow$ & MAE $\downarrow$ & Data \\
    \midrule
    \multicolumn{6}{@{}l}{\textit{Reported numbers}} \\
    \midrule
    Dlib~\cite{kazemi2014dlib}                   & 11.86 & 13.00 & 19.56 & 14.81 & R \\
    3DDFA~\cite{zhu2016300wlp}                   &  5.50 & 41.90 & 13.22 & 19.07 & M \\
    EVA-GCN~\cite{xin2021evagcn}                 &  4.01 &  4.78 &  2.98 &  3.92 & M \\
    HopeNet~\cite{ruiz2018hopenet}               &  4.81 &  6.61 &  3.27 &  4.89 & M \\
    QuatNet~\cite{hsu2018quatnet}                &  4.01 &  5.49 &  2.94 &  4.15 & M \\
    Liu \etal~\cite{liu2019facial}               &  4.12 &  5.61 &  3.15 &  4.29 & M \\
    FSA-Net~\cite{yang2019fsanet}                &  4.27 &  4.96 &  2.76 &  4.00 & M \\
    HPE~\cite{huang2020hpe}                      &  4.57 &  5.18 &  3.12 &  4.29 & M \\
    WHENet-V~\cite{zhou2020whenet}               &  3.60 &  4.10 &  2.73 &  3.48 & M \\
    RetinaFace~\cite{deng2020retinaface}         &  4.07 &  6.42 &  2.97 &  4.49 & R \\
    FDN~\cite{zhang2020fdn}                      &  4.52 &  4.70 &  2.56 &  3.93 & M \\
    MNN~\cite{valle2020mnn}                      &  3.98 &  4.61 &  2.39 &  3.66 & M \\
    TriNet~\cite{cao2021trinet}                  &  3.05 &  4.76 &  4.11 &  3.97 & M \\
    6DRepNet~\cite{hempel20226drepnet}           &  3.24 &  4.48 &  2.68 &  3.47 & M \\
    Cao \etal~\cite{cao2022unbiased}             &  4.21 &  3.52 &  3.10 &  3.61 & M \\
    TokenHPE~\cite{zhang2023tokenhpe}            &  3.95 &  4.51 &  2.71 &  3.72 & M \\
    Cobo \etal~\cite{cobo2024representation}     &  4.58 &  4.65 &  2.71 &  3.98 & M \\
    img2pose~\cite{albiero2021img2pose}          &  4.57 &  3.55 &  3.24 &  3.79 & M \\
    PerspNet~\cite{kao2023perspnet}              &  3.10 & \underline{3.37} & \underline{2.38} &  2.95 & R \\
    TRG~\cite{chun2024trg}                       & \underline{3.04} &  3.44 & \textbf{1.78} & \textbf{2.75} & M \\
    VGGT-HPE (Rel., ours)                        & \textbf{2.24} & \textbf{3.04} &  3.17 & \underline{2.82} & \textbf{S} \\
    \midrule
    \multicolumn{6}{@{}l}{\textit{Reproduced under shared MTCNN protocol}} \\
    \midrule
    6DRepNet~\cite{hempel20226drepnet}           & \underline{3.74} & \underline{4.95} & \textbf{3.04} & \underline{3.91} & M \\
    TokenHPE-v1~\cite{zhang2023tokenhpe}         &  5.57 &  6.23 &  3.79 &  5.20 & M \\
    TRG~\cite{chun2024trg}                       &  4.58 &  7.18 &  3.68 &  5.15 & M \\
    VGGT-HPE-Abs (ours)                          &  4.90 &  7.01 &  3.53 &  5.15 & \textbf{S} \\
    VGGT-HPE (Rel., ours)                        & \textbf{2.24} & \textbf{3.04} & \underline{3.17} & \textbf{2.82} & \textbf{S} \\
    \bottomrule
  \end{tabular}%
  }
\end{table}

\subsection{Cross-Domain Evaluation on BIWI}

Table~\ref{tab:biwi_cross_domain} presents the main cross-domain evaluation, split into two sections. The top section collects numbers reported in the original publications (following the compilation of~\cite{chun2024trg}); however, these results are not directly comparable, as each method uses its own face detector, crop strategy, and evaluation subset. To enable a fair comparison, the bottom section reproduces all methods under a shared evaluation protocol using MTCNN~\cite{zhang2016mtcnn} detection, following the widely adopted setup of 6DRepNet~\cite{hempel20226drepnet}. For TokenHPE, only the v1 checkpoint is publicly available, so we report results with that model. For TRG, we observed that the authors evaluate only on the subset of BIWI frames for which their preprocessing pipeline (based on FAN and MTCNN) successfully detects a face; since this filtering is not documented, we re-ran TRG under our shared protocol for consistency. For VGGT-HPE (Rel.), we use as a fixed anchor throughout each BIWI subject sequence the first frame of each recording, requiring essentially \textit{only a single ground-truth pose per subject} rather than per-frame annotations.

In the reported-numbers section, VGGT-HPE (Rel.) achieves the lowest yaw ($2.24^\circ$) and pitch ($3.04^\circ$) errors among all methods, and the second-lowest MAE ($2.82^\circ$), behind only TRG ($2.75^\circ$), while being the only method trained exclusively on synthetic data. Under the controlled shared protocol, VGGT-HPE (Rel.) achieves the lowest MAE ($2.82^\circ$) among the baselines.

The key takeaway emerges from comparing our two variants: VGGT-HPE-Abs, which shares the same backbone and synthetic training data but operates in absolute single-image mode, reaches only $5.15^\circ$, while switching to the relative formulation yields the best result in the table. This confirms that the advantage is structural rather than architectural. Absolute methods must internalize a canonical reference frame from the training data, making them sensitive to domain shift. The relative formulation sidesteps this by predicting a displacement between two explicitly provided views, removing the dependence on an implicit anchor.

\begin{figure*}[t]
  \centering
  \includegraphics[width=\textwidth]{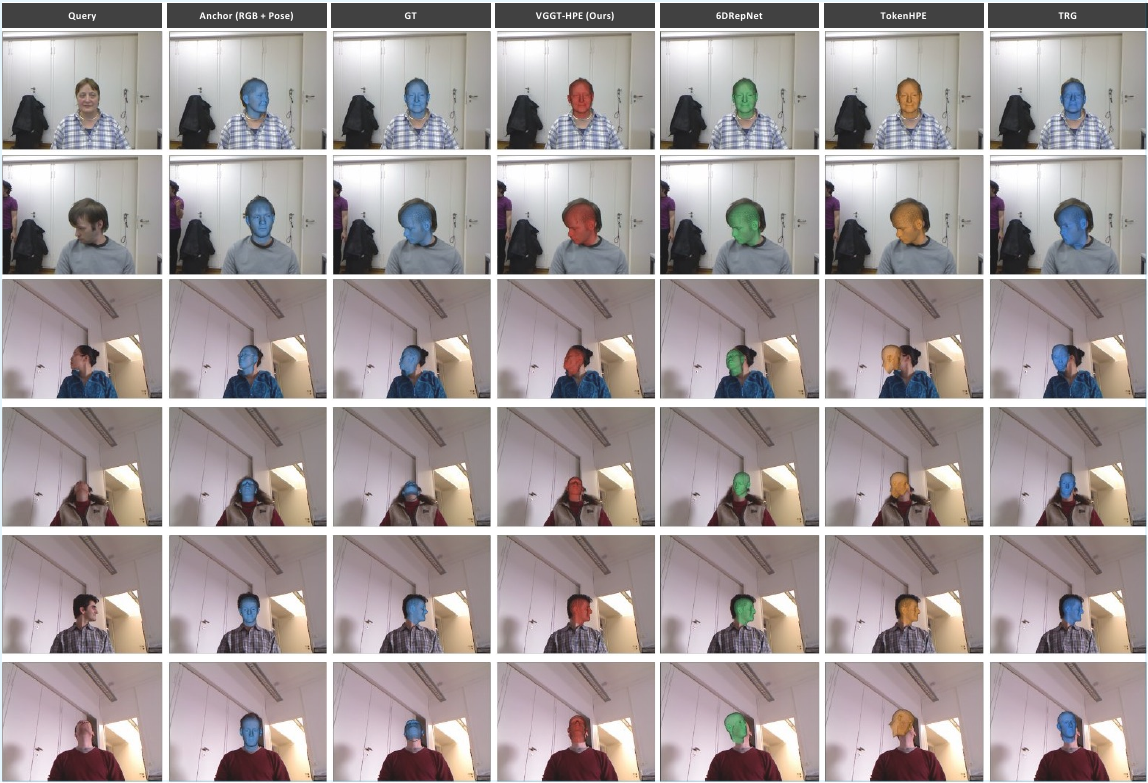}
  \caption{Qualitative results on BIWI. Each row shows a different subject. From left to right: the query frame, the anchor frame with its known pose overlay, the ground-truth pose, our prediction (VGGT-HPE), and three baselines (6DRepNet, TokenHPE, TRG). Our method produces pose estimates that are visually closer to the ground truth across a range of head orientations, including challenging near-profile and upward-looking poses where absolute methods exhibit larger deviations.}
  \label{fig:qualitative}
\end{figure*}

In Fig.~\ref{fig:qualitative} we also show qualitative results among various methods on the BIWI benchmark.

\begin{table}[t]
  \caption{BIWI hard benchmark (360 neutral-anchor / extreme-query pairs). Rotation errors in degrees (lower is better). Best in \textbf{bold}, second-best \underline{underlined}. ``Data'' indicates the training regime (S: synthetic-only, M: mixed) as defined in Table~\ref{tab:biwi_cross_domain}.}
  \label{tab:biwi_hardpairs}
  \centering
  \resizebox{\columnwidth}{!}{%
  \begin{tabular}{@{}lccccc@{}}
    \toprule
    Method & Yaw $\downarrow$ & Pitch $\downarrow$ & Roll $\downarrow$ & MAE $\downarrow$ & Data \\
    \midrule
    VGGT-HPE-Abs                                  & 40.74 & \underline{18.65} & 33.66 & 31.02 & \textbf{S} \\
    TokenHPE~\cite{zhang2023tokenhpe}              & 21.85 & 26.35 & 19.34 & 22.51 & M \\
    TRG~\cite{chun2024trg}                        & \underline{8.95} & 33.88 &  8.87 & 17.23 & M \\
    6DRepNet~\cite{hempel20226drepnet}             & 14.27 & 18.91 & \textbf{6.81} & \underline{13.33} & M \\
    VGGT-HPE (Rel., ours)                         & \textbf{3.81} & \textbf{15.87} & \underline{6.93} & \textbf{8.87} & \textbf{S} \\
    \bottomrule
  \end{tabular}%
  }
\end{table}

\subsection{Controlled Anchor Benchmarks: Easy and Hard Pairs}

To study how prediction difficulty scales with the anchor--target rotation gap, we construct two complementary benchmarks from BIWI, each with 360 pairs per subject. In both cases the anchor is near-neutral, so the absolute baselines-which ignore the anchor entirely-are evaluated directly on the target frames. In the hard benchmark these targets are extreme poses; in the easy one they are near-frontal. This lets us study how the different absolute and relative models behave in both easy and hard (extreme) poses.

\paragraph{Hard-pair benchmark.}
Each pair is formed by selecting a near-neutral anchor frame and an extreme-pose query frame, with a mean anchor--target rotation gap of approximately $70^\circ$. Results are shown in Table~\ref{tab:biwi_hardpairs}. The performance gap between VGGT-HPE (Rel.) and the absolute baselines widens substantially under these conditions: while 6DRepNet, the strongest absolute baseline, achieves $13.33^\circ$ MAE, VGGT-HPE (Rel.) reaches $8.87^\circ$, a $33\%$ relative improvement. The absolute variant of our model collapses entirely to $31.02^\circ$, confirming once more that the benefits stem from the relative formulation rather than the backbone.

\paragraph{Easy-pair benchmark.}
Here the situation is reversed: anchor and target are both near-neutral, with a mean rotation gap of only $3.82^\circ$. In this regime the absolute baselines are evaluated on their easiest samples --- target poses lie close to the implicit canonical frame that these methods internalize during training. Results are shown in Table~\ref{tab:biwi_easy}. Even in this favorable setting for absolute methods, VGGT-HPE (Rel.) achieves $0.96^\circ$ MAE, nearly three times better than 6DRepNet ($2.80^\circ$) and the absolute VGGT variant ($3.98^\circ$). This result highlights a key practical advantage of the relative formulation: when the anchor--target gap is small, the prediction problem becomes almost trivially easy.



\begin{table}[t]
  \caption{BIWI easy neutral-anchor benchmark (360 pairs; pair delta mean: $3.82^\circ$). Rotation errors in degrees (lower is better). Best in \textbf{bold}, second-best \underline{underlined}. ``Data'' indicates the training regime (S: synthetic-only, M: mixed).}
  \label{tab:biwi_easy}
  \centering
  \resizebox{\columnwidth}{!}{%
  \begin{tabular}{@{}lccccc@{}}
    \toprule
    Method & Yaw $\downarrow$ & Pitch $\downarrow$ & Roll $\downarrow$ & MAE $\downarrow$ & Data \\
    \midrule
    VGGT-HPE-Abs                                  &  5.62 & \underline{2.26} &  4.06 &  3.98 & \textbf{S} \\
    TokenHPE~\cite{zhang2023tokenhpe}              &  3.41 &  5.99 &  1.43 &  3.61 & M \\
    TRG~\cite{chun2024trg}                        &  3.59 &  4.24 &  2.52 &  3.45 & M \\
    6DRepNet~\cite{hempel20226drepnet}             & \underline{2.31} &  4.93 & \underline{1.14} & \underline{2.80} & M \\
    VGGT-HPE (Rel., ours)                         & \textbf{1.17} & \textbf{0.74} & \textbf{0.97} & \textbf{0.96} & \textbf{S} \\
    \bottomrule
  \end{tabular}%
  }
\end{table}

\subsection{Error Analysis: Relative Gap vs.\ Absolute Pose}

The easy- and hard-pair benchmarks test two ends of the difficulty range. Here we look at the full picture by binning results in $5^\circ$ increments along two axes: the anchor--query rotation gap and the absolute query orientation.

\begin{figure}[t]
  \centering
  \includegraphics[width=0.85\linewidth]{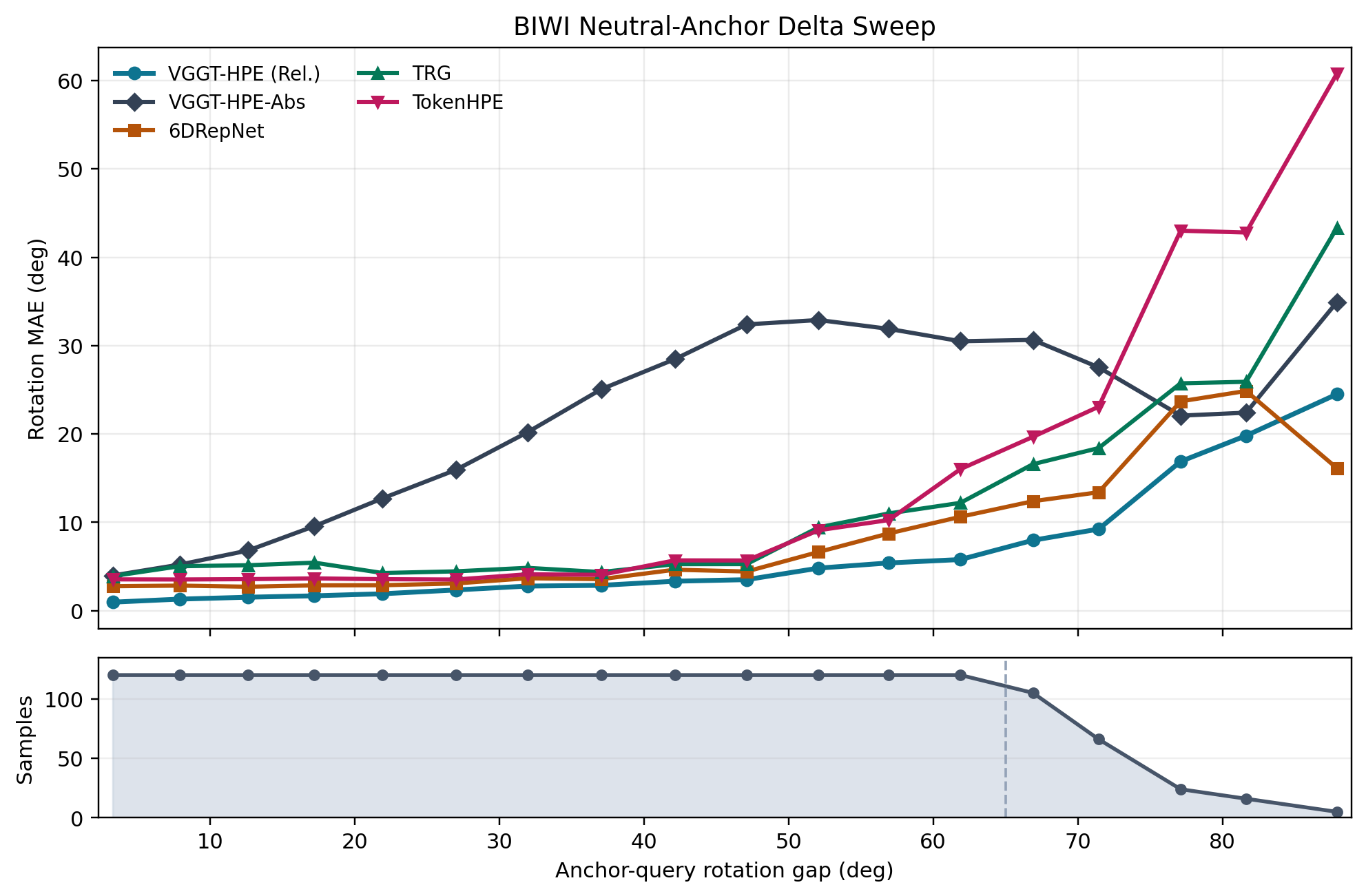}
  \caption{BIWI neutral-anchor evaluation as a function of anchor-query rotation gap. The upper plot reports rotation MAE, while the lower band shows the number of sampled pairs per bin. VGGT-HPE remains the strongest method across the full range.}
  \label{fig:biwi_neutral_anchor_delta}
\end{figure}

\paragraph{Scaling with the Anchor--Query Gap.}
In Figure~\ref{fig:biwi_neutral_anchor_delta} we bin anchor--target pairs by their ground-truth rotation gap in $5^\circ$ increments and plot the per-bin MAE for each method. Since for this test the anchor is always near-neutral, the gap directly reflects how far the query deviates from frontal. All methods exhibit increasing error as the gap grows, but VGGT-HPE (Rel.) starts from a lower baseline and rises more gradually than every absolute baseline, including our own absolute variant. This directly validates the hypothesis that relative head pose prediction is fundamentally easier, and that the advantage of the relative formulation grows with the magnitude of the displacement being estimated.

\begin{table}[t]
    \caption{Ablation study on synthetic validation data and BIWI (cross-domain). Rotation errors in degrees, sorted by decreasing MAE on BIWI. Best in \textbf{bold}, second-best \underline{underlined}. The absolute single-image variant (Abs.\ Single) fits the synthetic distribution best but transfers worst to BIWI, while the full relative model (VGGT-HPE) reverses this ranking --- evidence that the relative formulation transfers better across domains.}  \label{tab:ablation}
  \centering
  \resizebox{\columnwidth}{!}{%
  \begin{tabular}{@{}l cccc cccc@{}}
    \toprule
    & \multicolumn{4}{c}{Synthetic} & \multicolumn{4}{c}{BIWI} \\
    \cmidrule(lr){2-5} \cmidrule(lr){6-9}
    Variant & Yaw $\downarrow$ & Pitch $\downarrow$ & Roll $\downarrow$ & MAE $\downarrow$ & Yaw $\downarrow$ & Pitch $\downarrow$ & Roll $\downarrow$ & MAE $\downarrow$ \\
    \midrule
    \multicolumn{9}{@{}l}{\textit{Adaptation strategy}} \\
    \midrule
    Full finetune & 38.00 & 33.76 & 31.12 & 34.29 & 23.07 & 17.90 & 10.05 & 17.00 \\
    From scratch  &  9.21 & 15.07 & 14.05 & 12.78 &  7.71 &  8.12 &  6.94 &  7.59 \\
    Head-only     &  3.82 &  6.50 &  5.89 &  5.40 & 18.08 & 15.17 &  8.25 & 13.83 \\
    LoRA (ours)   &  \textbf{2.46} &  \textbf{4.65} &  \textbf{4.51} &  \textbf{3.87} &  \textbf{2.24} &  \textbf{3.04} &  \textbf{3.17} &  \textbf{2.82} \\
    \midrule
    \multicolumn{9}{@{}l}{\textit{Loss and formulation variants (all LoRA)}} \\
    \midrule
    Small-Gap     & 17.03 & 23.20 & 21.77 & 20.66 & 7.38  & 7.07  & 3.88  & 6.11 \\
    Abs.\ Pair    & 3.12  & 4.59  & 7.01  & 4.91  & 3.61  & 6.99  & 6.49  & 5.69 \\
    Abs.\ Single  & 2.44  & \underline{4.03} & \textbf{3.24} & \textbf{3.24} & 4.90  & 7.01  & 3.53  & 5.15 \\
    T-Aux, No FoV & 2.94  & 5.45  & 6.01  & 4.80  & 2.94  & 3.91  & 3.90  & 3.59 \\
    No FoV        & 2.39  & 4.13  & 4.19  & 3.57  & 2.64  & 4.05  & 3.61  & 3.43 \\
    T-Aux         & 2.43  & 4.37  & 4.48  & 3.76  & 2.76  & \underline{3.10} & 3.90 & 3.25 \\
    Geo Loss      & \underline{2.28} & \textbf{3.99} & \underline{3.76} & \underline{3.34} & 2.95  & 3.24  & 3.32  & 3.17 \\
    Rot.-Only     & 2.40  & 4.31  & 4.06  & 3.59  & \underline{2.58} & 3.48 & 3.32 & 3.12 \\
    Baseline      & \textbf{2.19} & 4.15 & 3.96 & 3.43 & 2.61  & 3.17  & \underline{3.31} & \underline{3.03} \\
    VGGT-HPE      & 2.46  & 4.65  & 4.51  & 3.87  & \textbf{2.24} & \textbf{3.04} & \textbf{3.17} & \textbf{2.82} \\
    \bottomrule
  \end{tabular}%
  }
\end{table}

\begin{figure}[t]
  \centering
  \includegraphics[width=0.85\linewidth]{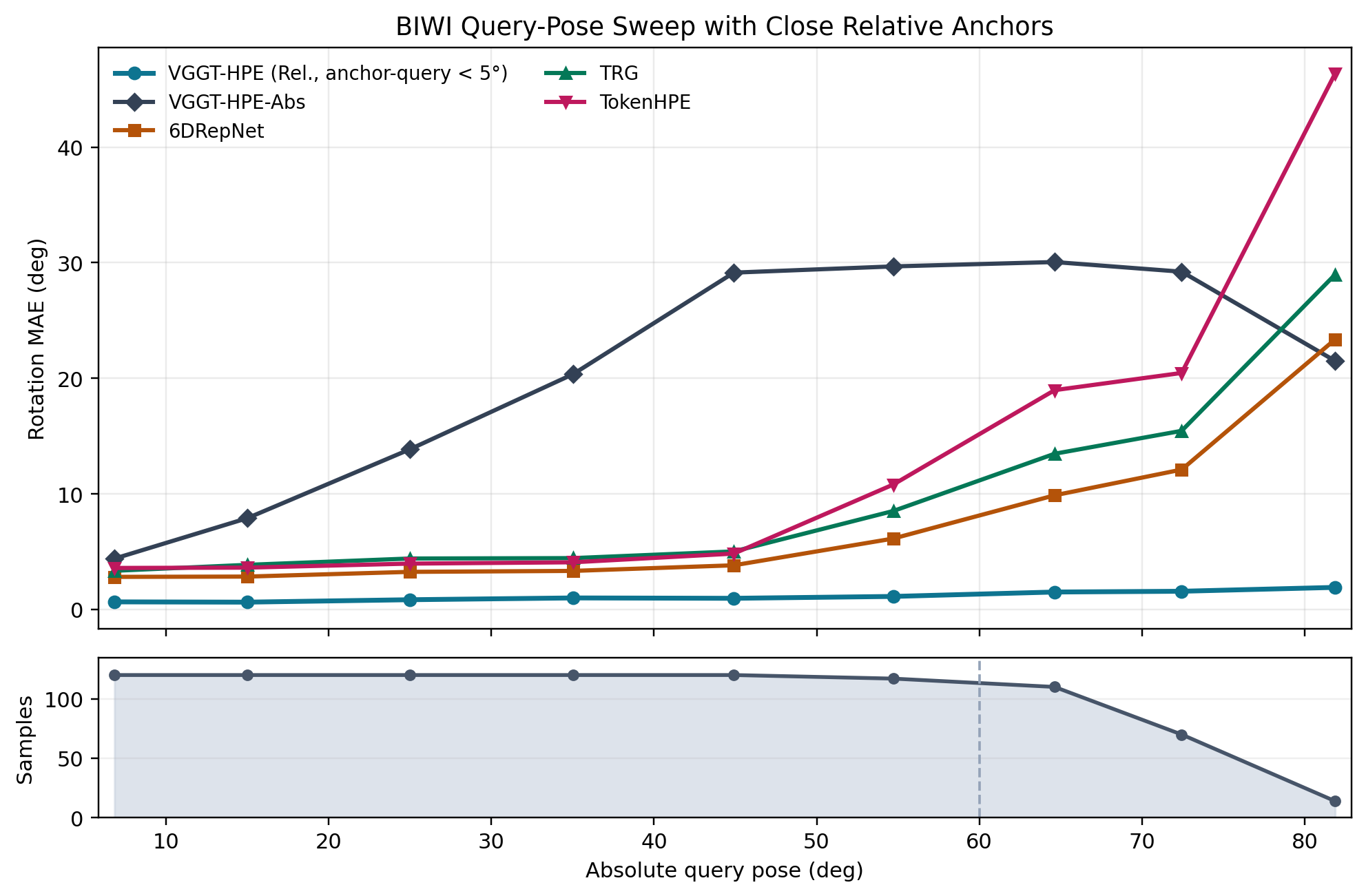}
  \caption{BIWI query-pose evaluation as a function of absolute query pose. For VGGT-HPE, each query is paired with a same-subject anchor whose anchor-query geodesic gap is below $5^\circ$. The upper plot reports rotation MAE, while the lower band shows the number of sampled pairs per bin.}
  \label{fig:biwi_query_pose}
\end{figure}

\paragraph{Robustness to Extreme Poses via Close Anchors.}
Absolute methods have no way to adapt at test time --- they always regress from a single crop relative to a fixed canonical frame learned during training, and performance drops as the target moves further from that reference. The relative formulation does not have this limitation: the anchor can be chosen to keep the prediction gap small, regardless of how extreme the target pose is. Figure~\ref{fig:biwi_query_pose} tests this directly. We sweep over absolute query pose but pair each query with a same-subject anchor within $5^\circ$ geodesic distance. The absolute baselines degrade beyond $60^\circ$, while VGGT-HPE (Rel.) stays below $2^\circ$ MAE across the full range --- the curve is nearly flat. This shows that what matters for relative prediction is the anchor--query gap, not the absolute pose of either view. With a close enough anchor, extreme poses are no harder than frontal ones.

\subsection{Ablation Studies}


Table~\ref{tab:ablation} presents ablation results on both the synthetic validation set and BIWI.  The top section compares adaptation strategies for the VGGT backbone. Full fine-tuning destroys the pretrained representations and performs worst on both datasets. Training from scratch (i.e., without pretrained weights) does better but still lags far behind LoRA on BIWI ($7.59^\circ$ vs.\ $2.82^\circ$). Head-only tuning is an interesting case: it achieves reasonable synthetic error ($5.40^\circ$) but collapses on BIWI ($13.83^\circ$), suggesting it overfits to the synthetic pose distribution without learning transferable features. LoRA strikes the best balance, preserving the pretrained geometric priors while adapting to the facial domain.

The bottom section ablates loss and formulation variants. The absolute single-image variant achieves the best synthetic MAE ($3.24^\circ$) but degrades to $5.15^\circ$ on BIWI, while VGGT-HPE reaches $2.82^\circ$ --- a clear sign that the relative formulation transfers better across domains. The Small-Gap variant performs worst on both datasets, confirming that sufficient anchor--query displacement is needed during training. Among the loss components, removing field-of-view prediction (No FoV; $3.43^\circ$) and translation supervision (T-Aux, No FoV; $3.59^\circ$) each hurt cross-domain performance. The geodesic loss and rotation-only variants are competitive, but the full combination in VGGT-HPE yields the best cross-domain result.

\subsection{Relaxing the Ground-Truth Anchor Assumption}
\label{sec:relaxing_anchor}
\begin{table}[t]
    \caption{Relaxing the ground-truth anchor assumption on the full BIWI set. The anchor pose is provided by an external absolute model instead of ground truth. Rotation errors in degrees (lower is better).}
    \label{tab:biwi_full_variantB_anchor_only}
  \centering
  \resizebox{\columnwidth}{!}{%
  \begin{tabular}{@{}lcccc@{}}
    \toprule
    Method & Yaw $\downarrow$ & Pitch $\downarrow$ & Roll $\downarrow$ & MAE $\downarrow$ \\
    \midrule
    VGGT-HPE (Rel., GT anchor)            & \textbf{2.24} & \textbf{3.04} & \textbf{3.17} & \textbf{2.82}  \\
    VGGT-HPE (Rel., VGGT-HPE-Abs anchor)  & 4.56 & \underline{5.28} & 3.42 & 4.42  \\
    VGGT-HPE (Rel., 6DRepNet anchor)      & \underline{2.89} & 6.05 & \underline{3.29} & \underline{4.08}  \\
    VGGT-HPE (Rel., TokenHPE anchor)      & 3.35 & 7.30 & 3.91 & 4.85  \\
    VGGT-HPE (Rel., TRG anchor)           & 4.02 & 12.03 & 7.13 & 7.73  \\
    \bottomrule
  \end{tabular}%
  }
\end{table}

\begin{table}[t]
    \caption{Relaxing the ground-truth anchor assumption on the hard BIWI subset (360 neutral-anchor / extreme-query pairs). The anchor pose is provided by an external absolute model instead of ground truth. Rotation errors in degrees (lower is better).}
      \label{tab:biwi_hard_variantB_anchor_only}
  \centering
  \resizebox{\columnwidth}{!}{%
  \begin{tabular}{@{}lcccc@{}}
    \toprule
    Method & Yaw $\downarrow$ & Pitch $\downarrow$ & Roll $\downarrow$ & MAE $\downarrow$  \\
    \midrule
    VGGT-HPE (Rel., GT anchor)            & \textbf{3.81} & \textbf{15.87} & \underline{6.93} & \textbf{8.87}  \\
    VGGT-HPE (Rel., VGGT-HPE-Abs anchor)  & 6.93 & \underline{17.36} & 8.68 & 10.99 \\
    VGGT-HPE (Rel., 6DRepNet anchor)      & \underline{4.81} & 19.36 & \textbf{6.62} & \underline{10.26} \\
    VGGT-HPE (Rel., TokenHPE anchor)      & 5.92 & 20.31 & 7.44 & 11.22 \\
    VGGT-HPE (Rel., TRG anchor)           & 7.07 & 21.82 & 17.00 & 15.30 \\
    \bottomrule
  \end{tabular}%
  }
\end{table}

A practical limitation of the relative formulation is that it requires a known anchor pose at test time. To understand how sensitive our method is to the anchor pose quality, we replace the ground-truth anchor pose with the prediction of an external absolute model. For each BIWI anchor--query pair, we estimate the anchor pose using one of VGGT-HPE-Abs, 6DRepNet, TokenHPE, or TRG, and use that prediction as the reference for VGGT-HPE (Rel.). The model then predicts the relative transformation with respect to the predicted anchor and composes it to recover an absolute query pose. Since any error in the anchor estimate propagates directly into the composed query pose as a constant offset, this setting is expected to increase the overall error proportionally to the accuracy of the anchor model. Results on the full BIWI set and the hard subset are shown in Tables~\ref{tab:biwi_full_variantB_anchor_only} and~\ref{tab:biwi_hard_variantB_anchor_only}, respectively.

Performance naturally degrades compared to using the ground-truth anchor, but the drop is moderate when the anchor estimator is reasonably accurate. With a 6DRepNet anchor, for instance, VGGT-HPE (Rel.) reaches $4.08^\circ$ on the full set --- worse than with ground truth ($2.82^\circ$), but still competitive with the absolute baselines in Table~\ref{tab:biwi_cross_domain}. On the hard subset the same trend holds: the relative pipeline with a 6DRepNet anchor ($10.26^\circ$) still outperforms every standalone absolute method except 6DRepNet itself. These results suggest that the ground-truth anchor requirement is not a hard constraint --- a reasonable absolute estimator can provide a sufficient reference frame, and the relative formulation remains beneficial even when the anchor is imperfect.

\section{Limitations}
\label{sec:limitations}
The relative formulation requires a known anchor pose at test time. As shown in Section~\ref{sec:relaxing_anchor}, this requirement can be partially relaxed by using the prediction of an absolute model as the anchor, but performance degrades with anchor quality. In sequential settings---such as video conferencing, sign language recognition, and driver monitoring---prior frames naturally serve as anchors, though using the model's own predictions auto-regressively can lead to error accumulation and drift over time. The flat error curve in Figure~\ref{fig:biwi_query_pose} relies on access to geometrically close ground-truth anchors; deploying the method in continuous temporal settings would require periodic recalibration, multi-anchor consensus, or external stabilization to prevent drift.

\section{Conclusions}
We presented VGGT-HPE, a relative head pose estimator that reframes the problem as relative pose prediction between two views. Built on the VGGT geometry foundation model, our method achieves the best results on BIWI among all compared methods, despite training only on synthetic data, while the same backbone in absolute mode ranks among the weakest. This contrast alone highlights the power of the relative formulation. Our controlled easy-pair and hard-pair benchmarks show that the advantage holds across the full difficulty spectrum and grows as the anchor--target gap increases. The error-vs-gap analysis directly validates the core thesis of this work: relative prediction is fundamentally easier than absolute regression. Moreover, unlike absolute methods whose implicit reference frame is fixed at training time, the relative formulation offers the flexibility to choose the anchor at test time, adapting the difficulty of the prediction to the application at hand.

\section*{Acknowledgements}

\begin{wrapfigure}{r}{0.27\columnwidth}
    \vspace{0.8\baselineskip}
    \centering
    \includegraphics[width=\linewidth]{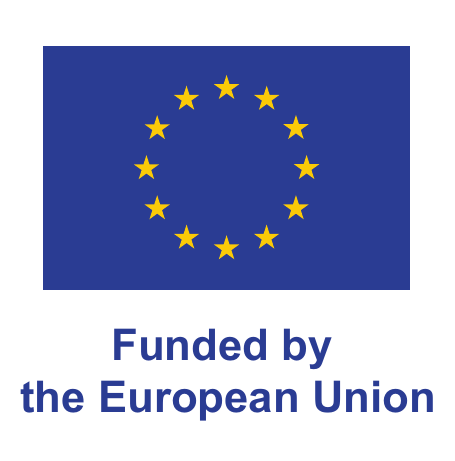}
    \vspace{-0.8\baselineskip}
\end{wrapfigure}

The work of V.~Vasileiou and K.~Daniilidis has been partially supported by project MIS 5154714 of the National Recovery and Resilience Plan Greece 2.0, funded by the European Union under the NextGenerationEU Program. The work of V.~Vasileiou, P.~P.~Filntisis, and P.~Maragos has also been partially funded by the European Union under Horizon Europe (grant No.~101136568 -- HERON). We acknowledge EuroHPC Joint Undertaking for awarding us access to Leonardo at CINECA, Italy, under project EHPC-DEV-2026D01-089.
{
    \small
    \bibliographystyle{ieeenat_fullname}
    \bibliography{main}

@String(CVPR= {IEEE Conf. Comput. Vis. Pattern Recog.})

@String(ECCV= {Eur. Conf. Comput. Vis.})

@String(ICIP = {IEEE Int. Conf. Image Process.})

@String(ICLR = {Int. Conf. Learn. Represent.})

@String(AAAI = {AAAI})

@String(CVPRW= {IEEE Conf. Comput. Vis. Pattern Recog. Worksh.})

@String(CVPR  = {CVPR})

@String(ECCV  = {ECCV})

@String(ICIP  = {ICIP})

@String(ICLR  = {ICLR})

@String(CVPRW= {CVPRW})

@inproceedings{wang2025vggt,
  title     = {{VGGT}: Visual Geometry Grounded Transformer},
  author    = {Wang, Jianyuan and Chen, Minghao and Karaev, Nikita and Vedaldi, Andrea and Rupprecht, Christian and Novotny, David},
  booktitle = {Proceedings of the IEEE/CVF Conference on Computer Vision and Pattern Recognition (CVPR)},
  pages     = {5294--5306},
  year      = {2025}
}

@inproceedings{li2017flame,
  title     = {Learning a Model of Facial Shape and Expression from {4D} Scans},
  author    = {Li, Tianye and Bolkart, Timo and Black, Michael J. and Li, Hao and Romero, Javier},
  booktitle = {ACM SIGGRAPH Asia},
  volume    = {36},
  number    = {6},
  year      = {2017}
}

@article{fanelli2013biwi,
  title   = {Random Forests for Real Time {3D} Face Analysis},
  author  = {Fanelli, Gabriele and Dantone, Matthias and Gall, Juergen and Fossati, Andrea and Van Gool, Luc},
  journal = {International Journal of Computer Vision},
  volume  = {101},
  pages   = {437--458},
  year    = {2013}
}

@inproceedings{hempel20226drepnet,
  title     = {{6D} Rotation Representation for Unconstrained Head Pose Estimation},
  author    = {Hempel, Thorsten and Abdelrahman, Ahmed A. and Al-Hamadi, Ayoub},
  booktitle = {IEEE International Conference on Image Processing (ICIP)},
  pages     = {2496--2500},
  year      = {2022}
}

@inproceedings{zhang2023tokenhpe,
  title     = {{TokenHPE}: Learning Orientation Tokens for Efficient Head Pose Estimation via Transformers},
  author    = {Zhang, Cheng and Liu, Hai and Deng, Yongjian and Xie, Bochen and Li, Youfu},
  booktitle = {Proceedings of the IEEE/CVF Conference on Computer Vision and Pattern Recognition (CVPR)},
  pages     = {8897--8906},
  year      = {2023}
}

@inproceedings{chun2024trg,
  title     = {{6DoF} Head Pose Estimation through Explicit Bidirectional Interaction with Face Geometry},
  author    = {Chun, Sungho and Chang, Ju Yong},
  booktitle = {European Conference on Computer Vision (ECCV)},
  year      = {2024}
}

@inproceedings{ruiz2018hopenet,
  title     = {Fine-Grained Head Pose Estimation Without Keypoints},
  author    = {Ruiz, Nataniel and Chong, Eunji and Rehg, James M.},
  booktitle = {IEEE/CVF Conference on Computer Vision and Pattern Recognition Workshops (CVPRW)},
  year      = {2018}
}

@inproceedings{yang2019fsanet,
  title     = {{FSA-Net}: Learning Fine-Grained Structure Aggregation for Head Pose Estimation from a Single Image},
  author    = {Yang, Tsun-Yi and Chen, Yi-Ting and Lin, Yin-Yu and Chuang, Yung-Yu},
  booktitle = {Proceedings of the IEEE/CVF Conference on Computer Vision and Pattern Recognition (CVPR)},
  pages     = {1087--1096},
  year      = {2019}
}

@article{zhou2020whenet,
  title   = {{WHENet}: Real-Time Fine-Grained Estimation for Wide Range Head Pose},
  author  = {Zhou, Yijun and Gregson, James},
  journal = {arXiv preprint arXiv:2005.10353},
  year    = {2020}
}

@inproceedings{cao2021trinet,
  title     = {A Vector-Based Representation to Enhance Head Pose Estimation},
  author    = {Cao, Zhiwen and Chu, Zongcheng and Liu, Dongfang and Chen, Yingjie},
  booktitle = {IEEE/CVF Winter Conference on Applications of Computer Vision (WACV)},
  year      = {2021}
}

@inproceedings{albiero2021img2pose,
  title     = {{img2pose}: Face Alignment and Detection via {6DoF}, Face Pose Estimation},
  author    = {Albiero, V{\'\i}tor and Chen, Xingyu and Yin, Xi and Pang, Guan and Hassner, Tal},
  booktitle = {Proceedings of the IEEE/CVF Conference on Computer Vision and Pattern Recognition (CVPR)},
  year      = {2021}
}

@article{kao2023perspnet,
  title   = {Toward {3D} Face Reconstruction in Perspective Projection: Estimating {6DoF} Face Pose from Monocular Image},
  author  = {Kao, Yueying and Pan, Bowen and Xu, Miao and Lyu, Jiangjing and Zhu, Xiangyu and Chang, Yanbo and Li, Xiaobo and Lei, Zhen},
  journal = {IEEE Transactions on Image Processing},
  volume  = {32},
  pages   = {3080--3091},
  year    = {2023}
}

@inproceedings{zhu2016300wlp,
  title     = {Face Alignment across Large Poses: A {3D} Solution},
  author    = {Zhu, Xiangyu and Lei, Zhen and Liu, Xiaoming and Shi, Hailin and Li, Stan Z.},
  booktitle = {Proceedings of the IEEE Conference on Computer Vision and Pattern Recognition (CVPR)},
  pages     = {146--155},
  year      = {2016}
}

@inproceedings{hu2022lora,
  title     = {{LoRA}: Low-Rank Adaptation of Large Language Models},
  author    = {Hu, Edward J. and Shen, Yelong and Wallis, Phillip and Allen-Zhu, Zeyuan and Li, Yuanzhi and Wang, Shanen and Wang, Lu and Chen, Weizhu},
  booktitle = {International Conference on Learning Representations (ICLR)},
  year      = {2022}
}

@article{zhang2016mtcnn,
  title   = {Joint Face Detection and Alignment Using Multitask Cascaded Convolutional Networks},
  author  = {Zhang, Kaipeng and Zhang, Zhanpeng and Li, Zhifeng and Qiao, Yu},
  journal = {IEEE Signal Processing Letters},
  volume  = {23},
  number  = {10},
  pages   = {1499--1503},
  year    = {2016}
}

@misc{blender2018,
  title        = {Blender -- a {3D} Modelling and Rendering Package},
  author       = {{Blender Online Community}},
  howpublished = {Blender Foundation},
  year         = {2018}
}

@inproceedings{filntisis2026mochi,
  title     = {{MOCHI}: Registration-Free Learnable Multi-View Capture of Faces in Dense Semantic Correspondence},
  author    = {Filntisis, Panagiotis P. and Retsinas, George and Dane\v{c}ek, Radek and Sklyarova, Vanessa and Maragos, Petros and Bolkart, Timo},
  booktitle = {Proceedings of the IEEE/CVF Conference on Computer Vision and Pattern Recognition (CVPR)},
  year      = {2026}
}

@inproceedings{sklyarova2024haar,
  title     = {Text-Conditioned Generative Model of {3D} Strand-Based Human Hairstyles},
  author    = {Sklyarova, Vanessa and Zakharov, Egor and Hilliges, Otmar and Black, Michael J. and Thies, Justus},
  booktitle = {Proceedings of the IEEE/CVF Conference on Computer Vision and Pattern Recognition (CVPR)},
  pages     = {4703--4712},
  year      = {2024}
}

@inproceedings{kazemi2014dlib,
  title     = {One Millisecond Face Alignment with an Ensemble of Regression Trees},
  author    = {Kazemi, Vahid and Sullivan, Josephine},
  booktitle = {Proceedings of the IEEE Conference on Computer Vision and Pattern Recognition (CVPR)},
  year      = {2014}
}

@inproceedings{xin2021evagcn,
  title     = {{EVA-GCN}: Head Pose Estimation Based on Graph Convolutional Networks},
  author    = {Xin, Miao and Mo, Shuangtao and Lin, Yaoyang},
  booktitle = {Proceedings of the IEEE/CVF Conference on Computer Vision and Pattern Recognition (CVPR)},
  year      = {2021}
}

@article{hsu2018quatnet,
  title   = {{QuatNet}: Quaternion-Based Head Pose Estimation with Multiregression Loss},
  author  = {Hsu, Hao-Wei and Wu, Ting-Yang and Wan, Shen and Wong, Wing Hung and Lee, Chen-Yi},
  journal = {IEEE Transactions on Multimedia},
  volume  = {21},
  number  = {4},
  pages   = {1035--1046},
  year    = {2018}
}

@inproceedings{liu2019facial,
  title     = {Facial Pose Estimation by Deep Learning from Label Distributions},
  author    = {Liu, Zhiwen and Chen, Zhihang and Bai, Jing and Li, Shanshan and Lian, Shiguo},
  booktitle = {IEEE/CVF International Conference on Computer Vision Workshops (ICCVW)},
  year      = {2019}
}

@inproceedings{huang2020hpe,
  title   = {Improving Head Pose Estimation Using Two-Stage Ensembles with Top-k Regression},
  author  = {Huang, Bin and Chen, Renwen and Xu, Wang and Zhou, Qinbang},
  journal = {Image and Vision Computing},
  volume  = {93},
  pages   = {103827},
  year    = {2020}
}

@inproceedings{deng2020retinaface,
  title     = {{RetinaFace}: Single-Shot Multi-Level Face Localisation in the Wild},
  author    = {Deng, Jiankang and Guo, Jia and Ververas, Evangelos and Kotsia, Irene and Zafeiriou, Stefanos},
  booktitle = {Proceedings of the IEEE/CVF Conference on Computer Vision and Pattern Recognition (CVPR)},
  year      = {2020}
}

@inproceedings{zhang2020fdn,
  title     = {{FDN}: Feature Decoupling Network for Head Pose Estimation},
  author    = {Zhang, Hao and Wang, Mingyi and Liu, Yonggenui and Yuan, Yi},
  booktitle = {AAAI Conference on Artificial Intelligence},
  year      = {2020}
}

@article{valle2020mnn,
  title   = {Multi-Task Head Pose Estimation In-the-Wild},
  author  = {Valle, Roberto and Buenaposada, Jos{\'e} M. and Baumela, Luis},
  journal = {IEEE Transactions on Pattern Analysis and Machine Intelligence},
  volume  = {43},
  number  = {8},
  pages   = {2874--2881},
  year    = {2020}
}

@inproceedings{cao2022unbiased,
  title     = {Towards Unbiased Label Distribution Learning for Facial Pose Estimation Using Anisotropic Spherical {G}aussian},
  author    = {Cao, Zhiwen and Liu, Dongfang and Wang, Qijun and Chen, Yingjie},
  booktitle = {ECCV}}

@article{murphy2009head,
  title={Head pose estimation in computer vision: A survey},
  author={Murphy-Chutorian, Erik and Trivedi, Mohan Manubhai},
  journal={IEEE Transactions on Pattern Analysis and Machine Intelligence},
  volume={31},
  number={4},
  pages={607--626},
  year={2009},
  publisher={IEEE}
}

@article{abate2022head,
  title={Head pose estimation: An extensive survey on recent techniques and applications},
  author={Abate, Andrea F. and Bisogni, Carmen and Castiglione, Arcangelo and Nappi, Michele},
  journal={Pattern Recognition},
  volume={127},
  pages={108591},
  year={2022},
  publisher={Elsevier}
}

@article{zhang2019mpiigaze,
  title={{MPIIGaze}: Real-world dataset and deep appearance-based gaze estimation},
  author={Zhang, Xucong and Sugano, Yusuke and Fritz, Mario and Bulling, Andreas},
  journal={IEEE Transactions on Pattern Analysis and Machine Intelligence},
  volume={41},
  number={1},
  pages={162--175},
  year={2019},
  publisher={IEEE}
}

@article{jha2022driver,
  title={Estimation of driver's gaze region from head pose using a single {RGB} camera},
  author={Jha, Sumit and Busso, Carlos},
  journal={IEEE Transactions on Intelligent Transportation Systems},
  volume={23},
  number={10},
  pages={17907--17918},
  year={2022},
  publisher={IEEE}
}

@article{koller2020quantitative,
  title={Quantitative survey of the state of the art in sign language recognition},
  author={Koller, Oscar},
  journal={arXiv preprint arXiv:2008.09918},
  year={2020}
}

@article{mavridis2015review,
  title={A review of verbal and non-verbal human--robot interactive communication},
  author={Mavridis, Nikolaos},
  journal={Robotics and Autonomous Systems},
  volume={63},
  pages={22--35},
  year={2015},
  publisher={Elsevier}
}

@article{marchand2016pose,
  title={Pose estimation for augmented reality: A hands-on survey},
  author={Marchand, Eric and Uchiyama, Hideaki and Spindler, Fabien},
  journal={IEEE Transactions on Visualization and Computer Graphics},
  volume={22},
  number={12},
  pages={2633--2651},
  year={2016},
  publisher={IEEE}
}

@inproceedings{wang2024dust3r,
  title={{DUSt3R}: Geometric 3D Vision Made Easy},
  author={Wang, Shuzhe and Leroy, Vincent and Cabon, Yohann and Chidlovskii, Boris and Revaud, J{\'e}rome},
  booktitle={Proceedings of the IEEE/CVF Conference on Computer Vision and Pattern Recognition},
  pages={20697--20709},
  year={2024}
}

@inproceedings{weinzaepfel2022croco,
  title={{CroCo}: Cross-view completion for 3d vision},
  author={Weinzaepfel, Philippe and Leroy, Vincent and Lucas, Thomas and Br{\'e}gier, Romain and Cabon, Yohann and Arora, Vaibhav and Antsfeld, Leonid and Chidlovskii, Boris and Rogez, Gr{\'e}gory and Revaud, J{\'e}rome},
  booktitle={Advances in Neural Information Processing Systems},
  volume={35},
  pages={17424--17438},
  year={2022}
}

@article{alsfasser2024mast3r,
  title={{MASt3R}: Matching and Stereo 3D Reconstruction},
  author={Leroy, Vincent and Cabon, Yohann and Revaud, J{\'e}rome},
  journal={arXiv preprint arXiv:2406.09756},
  year={2024}
}

@inproceedings{engel2014lsd,
  title={LSD-SLAM: Large-scale direct monocular SLAM},
  author={Engel, Jakob and Sch{\"o}ps, Thomas and Cremers, Daniel},
  booktitle={European conference on computer vision},
  pages={834--849},
  year={2014},
  organization={Springer}
}

@inproceedings{gu2017rnn,
  title={Dynamic facial analysis: From bayesian filtering to recurrent neural network},
  author={Gu, Jinwei and Yang, Xiaodong and De Mello, Shalini and Kautz, Jan},
  booktitle={Proceedings of the IEEE conference on computer vision and pattern recognition},
  pages={1548--1557},
  year={2017}
}

@inproceedings{wang2024dvmnet,
  title={{DROID-SLAM}: Deep Visual {SLAM} for Monocular, Stereo, and {RGB-D} Cameras},
  author={Teed, Zachary and Deng, Jia},
  booktitle={Advances in Neural Information Processing Systems},
  volume={34},
  pages={7166--7177},
  year={2021}
}

@inproceedings{piazza2021deep,
  title={TartanVO: A Generalizable Learning-Based VO},
  author={Wang, Wenshan and Zhu, Yaoyu and Wang, Xin and Zeng, Yuwei and Ding, Mingyu},
  booktitle={Conference on Robot Learning},
  pages={1761--1772},
  year={2021},
  organization={PMLR}
}

@article{btc2022temporal,
  title={Temporal Modeling and Structure Aggregation for Video Head Pose Estimation},
  author={Wang, Xinyu and others}, 
  journal={IEEE Transactions on Pattern Analysis and Machine Intelligence},
  year={2022}
}

@inproceedings{kuhn2023relative,
  title={Domain Adaptation for Head Pose Estimation Using Relative Pose Consistency},
  author={Kuhn, Lukas and M{\"u}ller, Markus and others},
  booktitle={Proceedings of the IEEE/CVF Winter Conference on Applications of Computer Vision (WACV)},
  year={2023}
}

@article{cobo2024representation,
  title={On the representation and methodology for wide and short range head pose estimation},
  author={Cobo, Alejandro and Valle, Roberto and Buenaposada, Jos{\'e} M and Baumela, Luis},
  journal={Pattern Recognition},
  volume={149},
  pages={110263},
  year={2024},
  publisher={Elsevier}
}

@article{kollias2021affect,
  title={Affect Analysis in-the-Wild: Valence-Arousal, Expressions, Action Units and a Unified Framework},
  author={Kollias, Dimitrios and Zafeiriou, Stefanos},
  journal={arXiv preprint arXiv:2103.15792},
  year={2021}
}
}


\end{document}


\maketitle
\begin{figure}[t]
  \centering
  \includegraphics[width=0.95\linewidth]{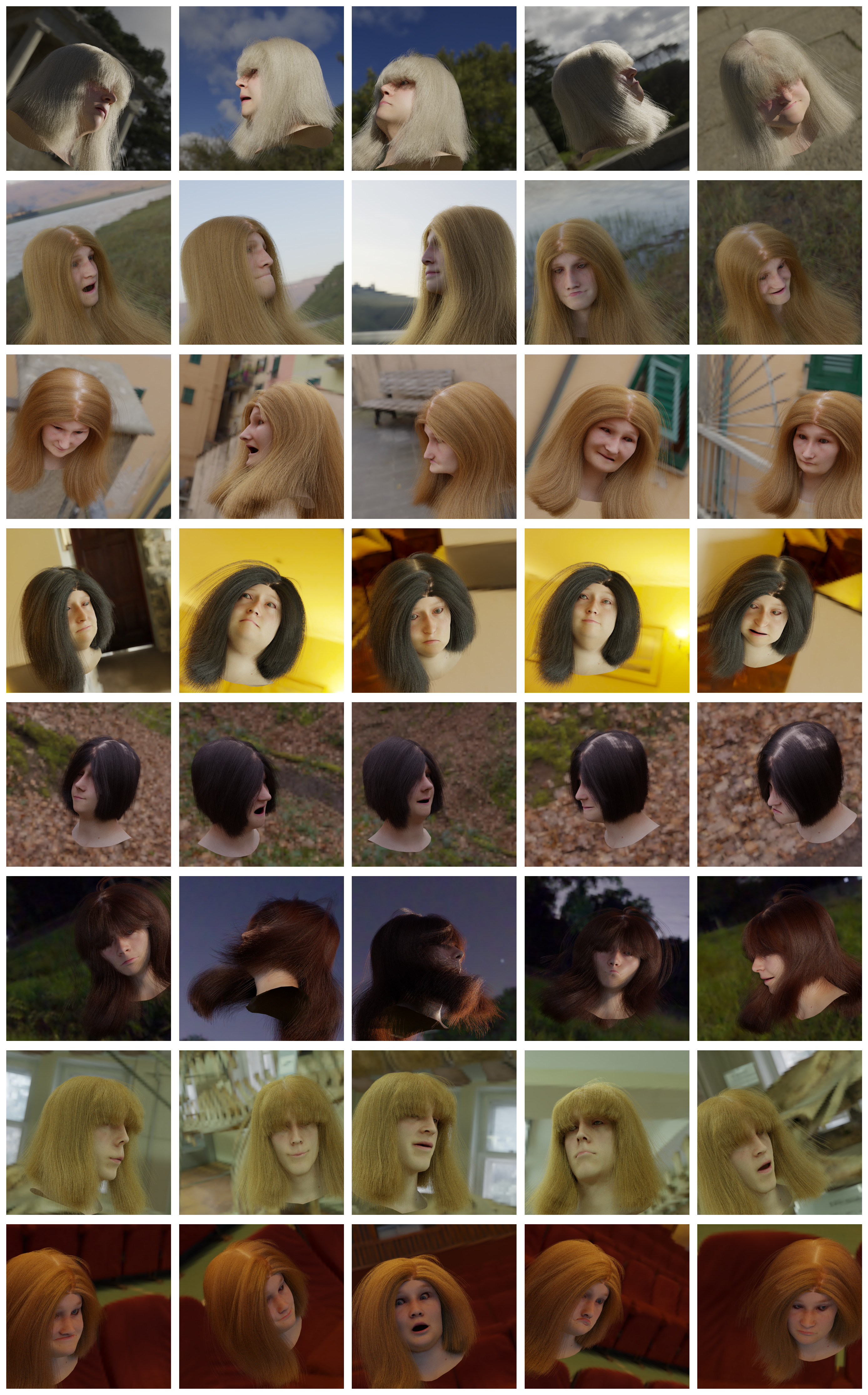}
  \label{fig:biwi_neutral_anchor_sweep}
\end{figure}
{
    \small
    \bibliographystyle{ieeenat_fullname}
    \bibliography{main}
}
